\documentclass[twoside]{article} 

\usepackage{aistats2017}

\usepackage{natbib}

\usepackage[utf8]{inputenc} 
\usepackage[T1]{fontenc}    
\usepackage{hyperref}       
\usepackage{url}            
\usepackage{booktabs}       
\usepackage{amsfonts}       
\usepackage{nicefrac}       
\usepackage{microtype}      
\usepackage{algorithmic}[1]
\usepackage{mathdots}
\usepackage{amsmath,graphicx,caption,color,epsfig,float,pbox,tabularx,wrapfig,mathrsfs,multirow,amsfonts,amsthm,times,subfigure,amssymb,mathrsfs,algorithm}
\usepackage{epstopdf}

\newcommand{\bfB}{\mathbf{B}}
\newcommand{\bfA}{\mathbf{A}}

\newcommand{\bfH}{\mathbf{H}}

\newcommand{\HB}{\mathbb{H}}
\newcommand{\bfI}{\mathbf{I}}

\newcommand{\bfy}{\mathbf{y}}
\newcommand{\bfx}{\mathbf{x}}

\newcommand{\RB}{\mathbb{R}}

\newtheorem{lemma}{Lemma}[section]

\newcommand{\EB}{\mathbb{E}}

\newcommand{\bfv}{\mathbf{v}}
\newcommand{\bfX}{\mathbf{X}}

\newcommand{\bfu}{\mathbf{u}}

\newcommand{\CL}{\mathcal{L}}

\begin{document}
	
	%
	
	%
	
	\twocolumn[
	
	\aistatstitle{Accelerating Stochastic Probabilistic Inference}
	\aistatsauthor{ Minta Liu, Suliang Bu }
	
	\aistatsaddress{ Amazon } ]
	
	\begin{abstract}
		Recently, Stochastic Variational Inference (SVI) has been increasingly attractive thanks to its ability to find good posterior approximations of probabilistic models. 
		It optimizes the variational objective with stochastic optimization, following noisy estimates of the natural gradient. 
		However, almost all the state-of-the-art SVI algorithms are based on first-order optimization algorithm and often suffer from poor convergence rate. 
		In this paper, we bridge the gap between second-order methods and stochastic variational inference by 
		proposing a second-order based stochastic variational inference approach. 
		In particular, firstly we derive the Hessian matrix of the variational objective. 
		Then we devise two numerical schemes to 
		implement second-order SVI efficiently. 
		Thorough empirical evaluations are investigated on both synthetic and real dataset to backup both the effectiveness and efficiency of the proposed approach. 
	\end{abstract}
	
	\section{Introduction}
	Performing large scale inference for complex models is a fundamental task in modern machine learning and statistical applications.
	Bayesian learning provides a probabilistic framework for inference that combines prior knowledge with observed data in a principled manner. 
	However, Bayesian computations are intractable in general case. 
	Thus one might resort to either Markov Chain Monte Carlo (MCMC)~\citep{robert2013monte} or variational Bayesian inference~\citep{jordan1999introduction}. 
	They are widely applied in probabilistic modelling \cite{chen2015convergence,chen2014stochastic}, and molecule generation~\cite{fu2021mimosa,fu2021probabilistic,huang2020deeppurpose}. 
	While MCMC provides unbiased estimates of Bayesian expectation, in practice designing MCMC algorithms that reliably converge to the desired posterior distribution is a notoriously difficult task especially in complex model. 
	On the other hand, variational Bayesian approach approximates the full posterior by attempting to minimize the Kullback-Leibler (KL)  divergence between the true posterior and a predefined distribution from a simple class of distribution on the same variables.
	Minimizing the KL divergence is equivalent to maximizing the familiar variational objective function. 
	Specifically, let $\Theta = \{\theta_1,\ldots,\theta_d\}$ denotes the parameter that we are interested and $\bfX$ the observed data.
	Variational methods approximate the intractable posterior distribution $p(\Theta\vert\bfX)$ with a $q$-distribution from a simple family of distribution.
	The $q$-distribution is characterized by variational parameter, denoted $\Gamma$. 
	For computational convenience, it is always assumed that the $q$-distribution is factorized. 
	That is, 
	\begin{equation}
		\label{eqn:factorize}
		\begin{aligned}
			q(\Theta\vert\Gamma) = \prod_{i=1}^{d} q_i(\theta_i\vert\gamma_i).
		\end{aligned}
	\end{equation}
	Accordingly, the variational parameter $\Gamma$ can be factorized as $\Gamma = \{\gamma_1,\ldots,\gamma_d\}$. 
	Factorized variational distribution has been proved to be efficient and effective. 
	Thus, in this paper, we pay our main attention to factorized variational distribution.
	
	The variational objective function (sometimes called variational lower bound) arises by bounding the marginal likelihood using the $q$-distribution, i.e., 
	\begin{equation}
		\label{eqn:variational_objective1}
		\begin{aligned}
			\CL(\Gamma) & = \int_{\Theta}^{} q(\Theta\vert\Gamma) \ln \frac{p(\bfX,\Theta)}{q(\Theta\vert\Gamma)} d\Theta \\
			& \leq  \ln \int_{\Theta}^{}p(\bfX,\Theta)d\Theta = \ln p(\bfX),\\	
		\end{aligned}
	\end{equation}
	where the inequality follows from the Jensen Inequality and $p(\bfX,\Theta) = p(\Theta)p(\bfX\vert\Theta)$ 
	Then the optimization problem can be formulated as
	\begin{equation}
		\label{eqn:variational_objective}
		\begin{aligned}
			\underset{\Gamma}{\arg\max}\  \CL(\Gamma)  = &  \EB_{q(\Theta\vert\Gamma)}[\ln p(\Theta\vert\bfX)] + \mathbb{H}(q(\Theta\vert\Gamma))\\
			&  + \ln p(\bfX), \\
		\end{aligned}
	\end{equation}
	where $\mathbb{H}(q(\Theta\vert\Gamma)) = - \int_{\Theta}^{}q(\Theta\vert\Gamma)\ln q(\Theta\vert\Gamma)d\Theta$, is a function of $\Gamma$ ($\Theta$ is integrated out). 
	Since $q$ comes from a simple family of distribution, say, exponential family,  $\mathbb{H}(q(\Theta\vert\Gamma))$ usually exhibits tractable form and differentiable w.r.t. the variational parameter.
	The last term (i.e., $\ln p(\bfX)$, logirithm of normalizing constant) on RHS, independent of $\Gamma$, can be omitted. 
	It is obverved that 
	$\CL(\Gamma) - \ln p(\bfX) = - \text{KL}(q(\Theta\vert\Gamma)\Vert p(\Theta\vert\bfX))$.
	Thus, maximizing $\CL$ with respect to $\Gamma$ is equivalent to minimizing the KL divergence between $q(\Theta\vert\Gamma)$ and $p(\Theta\vert\bfX)$.
	
	Unfortunately, Problem~\eqref{eqn:variational_objective} is not a typical optimization problem owing to the fact that both the variational objective and its gradient are intractable in general case, apart from simple cases involving some conjugate models. 
	To address this issue, \cite{paisley2012variational,ranganath2014black} presented a method to approximate the gradient of the variational lower bound. 
	In particular, they showed that the gradient of variational objective $\CL$ can be written into the form
	\begin{equation}
		\label{eqn:first_order_gradient}
		\begin{aligned}
			\nabla_{\Gamma}\CL(\Gamma) = &  \EB_{q(\Theta\vert\Gamma)}[ \nabla_\Gamma \log  q(\Theta\vert\Gamma) \ln p(\Theta, \bfX)]\\
			& + \nabla_{\Gamma} \mathbb{H}(q(\Theta\vert\Gamma)),	\\
		\end{aligned}
	\end{equation}
	where the second term on RHS owns closed form while Monte Carlo integration is adopted to approximate the first term, given as 
	\begin{equation}
		\label{eqn:first_order_gradient_2}
		\begin{aligned}
			\widehat{\nabla_{\Gamma}\CL(\Gamma)} = & \sum_{i=1}^{T} \nabla_\Gamma \log  q(\Theta^{(i)}\vert\Gamma) \ln p(\Theta^{(i)}\vert x)\\
			& + \nabla_{\Gamma} \mathbb{H}(q(\Theta\vert\Gamma)),\\  \text{where}\ \ 
			&  \Theta^{(1)},\ldots, \Theta^{(T)}\overset{\text{i.i.d.}}{\sim} q(\cdot\vert\Gamma).  \\ 
		\end{aligned}
	\end{equation}
	
Since the estimator is unbiased for the true gradient, 
a series of off-the-shelf tools in stochastic optimization can be adapted in this scenario~\citep{ranganath2013adaptive,wang2013variance}.
These methods are all based on first-order optimization algorithms.
	
	On the other hand, second-order methods, exploiting curvature information of the objective function, are acknowledged to enjoy faster per-iteration convergence and are well-studies in numerical optimization  literature~\citep{nocedal2006numerical}. 
	However, they have been much less explored in the context of stochastic variational inference owing to the absence of second-order information. 
	In this paper, we fill this blank and bridge the gap between second-order methods and stochastic variational inference. 
	In particular, we first derive the Hessian matrix of variational lower bound.  
	Then we provide two numerical schemes to implement second-order SVI efficiently.  
	Furthermore, empirical results are also satisfactory.

	The remainder of the paper is organized as follows:
	Section~\ref{sec:sg-mcmc} briefly introduces Newton's methods and its variants as preliminaries. 
	In Section~\ref{sec:method}, second-order SVI is demonstrated elaborately.

	\section{Preliminaries---Newton's Method} 
	\label{sec:sg-mcmc}
	In this section, we provide a concise description for Newton's method. 
	In the context of numerical optimization, given some differentiable function $f(\bfx):\RB^d\xrightarrow{}\RB$, the task is either minimize or maximize it by altering $\bfx$. 
	Among numerous optimization methods, the most commonly used method is steepest descent or gradient descent.
	Given current point $\bfx$, it proposes a new point $\bfx'$ via 
	\begin{equation}
		\label{eqn:gradient_descent}
		\begin{aligned}
			\bfx{'} = \bfx - \epsilon \nabla_{\bfx} f(\bfx),
		\end{aligned}
	\end{equation}
	where $\epsilon$ represents the learning rate, a positive scalar determining the size of the step. 
	Numerous variants are raised based on the updating rule in Equation~\eqref{eqn:gradient_descent}. 
	Optimization algorithms that use only gradients are called first-order optimization algorithms. 
	
	Accordingly, algorithms that use the Hessian matrix (defined later) are called second-order algorithms. 
	Newton's method, the mainstream second-order method, is also known as the Newton–Raphson method. 
	It is based on using a second-order Taylor series expansion to approximate $f(\bfx)$ near some point $\bfx^{(0)}$:
	\begin{equation*}
		\begin{aligned}
			f(\bfx) \approx & f(\bfx^{(0)}) + (\bfx - \bfx^{(0)})^T \nabla_{\bfx}f(\bfx^{(0)})\\
			& + \frac{1}{2}(\bfx - \bfx^{(0)})^T\bfH(f)(\bfx^{(0)})^{-1} (\bfx - \bfx^{(0)}), \\
		\end{aligned}
	\end{equation*}
	
	where $\bfH(f)(\bfx)\in\RB^{d\times d}$, referred to as the Hessian matrix of $f$, is defined such that 
	\begin{equation}
		\label{eqn:hessian2}
		\begin{aligned}
			\bfH(f)(\bfx)_{i,j} = \frac{\partial^2}{\partial \bfx_i \partial \bfx_j} f(\bfx) = \frac{\partial^2}{\partial\bfx_j\partial\bfx_i}f(\bfx). 
		\end{aligned}
	\end{equation}
	Equivalently, the Hessian is the Jacobian of the gradient. 
	Optimization algorithms such as Newton's method that use the Hessian matrix are called second-order optmization algorithms.
	If we solve for the critical point of this function, we obtain:
	\begin{equation}
		\label{eqn:newton_update}
		\begin{aligned}
			\bfx' = \bfx^{(0)} - \bfH(f) (\bfx^{(0)})^{-1} \nabla_{\bfx} f(\bfx^{(0)}).
		\end{aligned}
	\end{equation}
	It can be shown that the update rule described in Equation~\ref{eqn:newton_update} eventually reaches quadratic convergence under reasonable assumptions. 
	But it usually suffers from the prohibitively high per-iteration cost owing to the computation of the second-order information, especially the inversion of the Hessian matrix $\bfH(f) (\bfx^{(0)})$. 
	
	To handle this problem, a number of quasi-Newton methods were raised~\citep{nocedal2006numerical}, which require only the gradient of the objective function to approximate Hessian matrix. 
	The most popular quasi-Newton method is the BFGS method and L-BFGS ( limited-memory variant). 
	Worth to mention that \cite{fan2015fast,fu2020alpha} argued that L-BFGS works in the context of stochastic variational inference.

\section{Second-Order Stochastic Variational Inference}
\label{sec:method}	
In this section, we explore the possibility of marrying the second-order method with stochastic variational inference. 
First, we derive the Hessian matrix of $\CL(\Gamma)$.
Then the efficient implementation of second-order SGVI is studied and two schemes are developed. 
	
\subsection{Hessian matrix}	
The Hessian matrix of objective function plays the critical role in second-order methods. 
To derive the Hessian matrix of $\nabla^2_{\Gamma}\CL(\Gamma)$, 
we return to the gradient of variational objective $\nabla_{\Gamma}\CL(\Gamma)$ governed in Equation~\eqref{eqn:first_order_gradient}, 
whose derivation is given as 
\begin{equation}
	\begin{aligned}
\nabla_{\Gamma}\CL(\Gamma) = & \nabla_{\Gamma}\EB_{q}[\ln p(\Theta\vert\bfX)] + \nabla_{\Gamma} \mathbb{H}(q(\Theta\vert\Gamma))\\
= & \nabla_\Gamma \int_{\Theta}^{} q(\Theta\vert\Gamma) \ln p(\Theta\vert x)d\Theta + \nabla_{\Gamma} \mathbb{H}(q(\Theta\vert\Gamma)) \\
= & \int_{\Theta}^{} \nabla_\Gamma q(\theta\vert\Gamma) \ln p(\theta\vert x)d\Theta + \nabla_{\Gamma} \mathbb{H}(q(\Theta\vert\Gamma)) \\
= & \int_{\Theta}^{} q(\Theta\vert\Gamma) \nabla_\Gamma \log  q(\Theta\vert\Gamma) \ln p(\Theta\vert x)d\Theta\\
& + \nabla_{\Gamma} \mathbb{H}(q(\Theta\vert\Gamma))\\
= &  \EB_{q(\Theta\vert\Gamma)}[ \nabla_\Gamma \log  q(\Theta\vert\Gamma) \ln p(\Theta\vert x)]\\
& + \nabla_{\Gamma} \mathbb{H}(q(\Theta\vert\Gamma)),\\
	\end{aligned}
\end{equation}
where the fourth equality follows from the identity $\nabla_\Gamma q(\theta\vert\Gamma) = q(\Theta\vert\Gamma) \nabla_\Gamma \log q(\Theta\vert\Gamma)$, which is also known as log-derivative trick. 
In the above equation, the integration $ \int_{\Theta}^{} \nabla_\Gamma q(\theta\vert\Gamma) \ln p(\theta\vert x)d\Theta$ (the first term in third line) is computationally intractable generally while the integration $\int_{\Theta}^{} q(\Theta\vert\Gamma) \nabla_\Gamma \log  q(\Theta\vert\Gamma) \ln p(\Theta\vert x)d\Theta$ can be estimated by Monte Carlo method, as mentioned in Equation~\ref{eqn:first_order_gradient_2}. 
Based on this, we can derive the Hessian matrix $\nabla^2_{\Gamma}\CL(\Theta)$ via taking derivative to $\nabla_{\Gamma}\CL(\Gamma)$
	
\begin{equation*}
	\begin{aligned}
			& \nabla^2_{\Gamma}\CL(\Gamma) =  \nabla_{\Gamma}\{ \EB_{q(\Theta\vert\Gamma)}[ \nabla_\Gamma \log  q(\Theta\vert\Gamma) \ln p(\Theta,\bfX)]\\
			& + \nabla_\Gamma \HB(q(\Theta\vert\Gamma))\}\\
			= &  \nabla_\Gamma[\int_{\Theta}^{}q(\Theta\vert\Gamma)\nabla_\Gamma \log  q(\Theta\vert\Gamma) \ln p(\Theta,\bfX)]\\
			& + \nabla^2_\Gamma  \HB(q(\Theta\vert\Gamma))\\
			= & \int_\Theta \nabla_\Gamma q(\Theta\vert\Gamma) [\nabla_\Gamma \log  q(\Theta\vert\Gamma)]^T \ln p(\Theta,\bfX)d\Theta\\
			& + \int_{\Theta}^{} q(\Theta\vert\Gamma) \nabla^2_\Gamma \log  q(\Theta\vert\Gamma) \ln p(\Theta,\bfX)d\Theta\\
			&  + \nabla^2_\Gamma \HB(q(\Theta\vert\Gamma)) \\
			= & \int_\Theta q(\Theta\vert\Gamma) \nabla_\Gamma \log q(\Theta\vert\Gamma) [\nabla_\Gamma \log  q(\Theta\vert\Gamma)]^T  \ln p(\Theta,\bfX)d\Theta \\ & 
			+ \int_{\Theta}^{} q(\Theta\vert\Gamma) \nabla^2_\Gamma \log  q(\Theta\vert\Gamma) \ln p(\Theta,\bfX)d\Theta\\
			& + \nabla^2_\Gamma \HB(q(\Theta\vert\Gamma))\\
	\end{aligned}
\end{equation*}
	
\begin{equation}
	\label{eqn:hessian}
	\begin{aligned}
			= & \EB_{q(\Theta\vert\Gamma)} \{[\nabla_\Gamma \log q(\Theta\vert\Gamma)] [\nabla_\Gamma \log  q(\Theta\vert\Gamma)]^T \ln p(\Theta,\bfX)\\
			& +  \nabla^2_\Gamma \log  q(\Theta\vert\Gamma)\ln p(\Theta\vert x)\} + \nabla^2_\Gamma \HB(q(\Theta\vert\Gamma)),\\
	\end{aligned}
\end{equation}	
where the fourth equality (the eighth line) employs the log-derivative trick again, borrowing the idea from~\cite{paisley2012variational,ranganath2014black}. 
Worth to mention that for factorized variational distribution $q(\Theta\vert\Gamma)$ satisfying that $q(\Theta\vert\Gamma) = \prod_{i=1}^{d}q_i(\theta_i\vert\gamma_i)$, the matrix $\nabla^2_\Gamma \log  q(\Theta\vert\Gamma)$ is diagonal or block diagonal\footnote{The size of block is equal to the dimension of $\gamma_i$. if $\gamma_1,\ldots,\gamma_d\in \RB$, then the matrix reduces to diagonal.}. That is, 
\begin{equation}
		\label{eqn:diagonal_1}
		\begin{aligned}
			& \nabla^2_\Gamma \log  q(\Theta\vert\Gamma) =  \nabla^2_{\gamma_1,\ldots,\gamma_d} \sum_{i=1}^{d} \log q_i(\theta_i\vert\gamma_i) \\
			= & \begin{pmatrix} \nabla^2_{\gamma_1} \log q_1(\theta_1\vert\gamma_1) & & \\ & \ddots & \\ & & \nabla^2_{\gamma_d} \log q_d(\theta_d\vert\gamma_d) \end{pmatrix}.
		\end{aligned}
\end{equation}

	Similarly, we know that the last term $\nabla^2_\Gamma \HB(q(\Theta\vert\Gamma))$ in RHS of Equation~\eqref{eqn:hessian}
	can be simplified as
	\begin{equation}
		\label{eqn:diagonal_2}
		\begin{aligned}
			&\nabla^2_\Gamma \HB(q(\Theta\vert\Gamma)) =  \nabla^2_\Gamma(\sum_{i=1}^{d}\HB(q_i(\theta_i \vert \gamma_i)))\\
			= & \begin{pmatrix} \nabla^2_{\gamma_1} \HB(q_1(\theta_1 \vert \gamma_1)) & & \\ & \ddots & \\ & & \nabla^2_{\gamma_d} \HB(q_d(\theta_d \vert \gamma_d)) \end{pmatrix},\\
		\end{aligned}
	\end{equation}
where $\HB(q_i(\theta_i \vert \gamma_i))$ is a function of $\gamma_i$. 
Thus $\nabla^2_\Gamma \HB(q(\Theta\vert\Gamma))$ is also diagonal or block diagonal.
These structures would be exploited to cut down the complexity later. 
Then we discuss the implementation of second-order SVI. 
	
	The most straightforward approach is to stochastically approximate the Hessian matrix $\nabla^2_{\Gamma}\CL(\Theta)$ using Monte Carlo integration, 
	i.e.,
	\begin{equation}
		\label{eqn:monte_carlo_integration}
		\begin{aligned}
			\widehat{\nabla^2_{\Gamma}{\CL}(\Theta)} = & \frac{1}{S} \sum_{i=1}^{S}\{ [\nabla_\Gamma \log q(\Theta^{(i)}\vert\Gamma)] [\nabla_\Gamma \log  q(\Theta^{(i)}\vert\Gamma)]^T\\
			& \ln p(\Theta^{(i)}\vert x)
			+  \nabla^2_\Gamma \log  q(\Theta^{(i)}\vert\Gamma)\ln p(\Theta^{(i)}\vert x)\}\\
			& + \nabla^2_\Gamma \HB(q(\Theta\vert\Gamma)),\\
\text{where}\ \ & \Theta^{(1)},\ldots, \Theta^{(S)}	\overset{\text{i.i.d.}}{\sim}q(\Theta\vert\Gamma). \\  		
		\end{aligned}
	\end{equation}
	We can therefore replace $\nabla^2_{\Gamma}\CL(\Theta)$ with the unbiased stochastic approximation of the Hessian matrix in Equation~\eqref{eqn:hessian}. 
	Combining the update rule in Newton's method described in Equation~\eqref{eqn:newton_update}, the algorithm is shown in Algorithm~\ref{alg:0}.
	
	\begin{algorithm}[]
		\caption{A straightforward approach to implement second-order SGVI}
		\label{alg:0}
		\begin{algorithmic}[1]
			\REQUIRE initial value $\Gamma_0$, $S$, $T$, 
			\ENSURE final value of variational parameter $\Gamma_t$
			\FOR {$t = 1,2, \ldots$}
			\STATE Estimate gradient  $\widehat{\nabla_{\Gamma}\CL(\Gamma_{t-1})}$ according to Equation~\eqref{eqn:first_order_gradient_2}.\ \ \ \ \ \ \ \ \ \ \ \ \# $O(d)$
			\STATE Estimate Hessian matrix $\widehat{\nabla^2_{\Gamma}{\CL}(\Theta)}$ according to Equation~\eqref{eqn:monte_carlo_integration}.\ \ \ \ \ \ \ \ \ \ \ \ \# $O(d^2)$ 
			\STATE Update the variational parameter via $\Gamma_t = \Gamma_{t-1} -  (\widehat{\nabla^2_{\Gamma}{\CL}(\Theta)})^{-1}\widehat{\nabla_{\Gamma}\CL(\Gamma_{t-1})}$. \ \ \ \ \ \ \ \ \ \ \ \ \# $O(d^3)$
			\IF {Convergence condition is met}
			\STATE break.
			\ENDIF
			\ENDFOR
		\end{algorithmic}
	\end{algorithm} 
The computational complexity for each step is listed.
We find that per-iteration cost is prohibitively large due to the matrix inversion operation to the estimated Hessian matrix $\widehat{\nabla^2_{\Gamma}{\CL}(\Theta)}$, 
which costs $O(d^3)$ computation, as shown in Step 4 of Algorithm~\ref{alg:0}. 
	Thus we attempt to circumvent this operation in the follows. 
	
	\subsection{Scheme I}
	To do this, we first rearrange Equation~\eqref{eqn:monte_carlo_integration} as
\begin{equation}
	\label{eqn:monte_carlo_integration2}
	\begin{aligned}
			& \widehat{\nabla^2_{\Gamma}{\CL}(\Theta)}\\
			& = \underbrace{\frac{1}{S} \sum_{i=1}^{S}\{ \nabla^2_\Gamma \log  q(\Theta^{(i)}\vert\Gamma)\ln p(\Theta^{(i)}, \bfX)\}
				+ \nabla^2_\Gamma \HB(\Gamma)}_{\mathcal{A}}\\
			& + \underbrace{\sum_{i=1}^{S}\{ [c_i\nabla_\Gamma \log q(\Theta^{(i)}\vert\Gamma)] [c_i\nabla_\Gamma \log  q(\Theta^{(i)}\vert\Gamma)]^T}_{\mathcal{B}},\\
		& \text{where}\ \ \ c_i = \sqrt{\frac{\ln p(\Theta^{(i)},\bfX)}{S}}. 
	\end{aligned}
\end{equation}
We find that $\widehat{\nabla^2_{\Gamma}{\CL}(\Theta)}$, the estimated Hessian matrix, can be expressed as the sum of a diagonal (or block diagonal) matrix \footnote{corresponding to the term $\mathcal{A}$ in above equation, according to Equation~\eqref{eqn:diagonal_1} and ~\eqref{eqn:diagonal_2}.} and rank-$S$ correction \footnote{corresponding to the term $\mathcal{B}$}. 
	Then we demonstrate the celebrated Sherman–Morrison formula as follows. 
\begin{lemma}
Suppose $\bfA\in\RB^{d\times d}$ is an invertible square matrix and $\bfu, \bfv\in \RB^{d}$ are $d$-dimensional vector. 
Suppose furthermore that $1+v^T\bfA^{-1}u\neq 0 $.
Then the Sherman–Morrison formula states that
\begin{equation}
	\label{eqn:sherman}
	\begin{aligned}
(\bfA + \bfu\bfv^T)^{-1} = \bfA^{-1} - \frac{\bfA^{-1}\bfu\bfv^T\bfA^{-1} }{ 1 +\bfv^T\bfA^{-1}\bfu}.
	\end{aligned}
	\end{equation}
\end{lemma}
Note that the computation of Equation~\eqref{eqn:sherman} only involves matrix-vector product and requires $O(d^2)$ computation.
Thus we can iteratively use the Sherman–Morrison formula to compute the inversion of $\widehat{\nabla^2_{\Gamma}{\CL}(\Theta)}$ described in Equation~\ref{eqn:monte_carlo_integration}.
Each iteration requires $O(d^2)$ computations.
However, this strategy will be useful only in the case where $S\ll d$.
	
Then we focus on a more general case. 
In particular, our target is to compute 
$(\widehat{\nabla^2_{\Gamma}{\CL}(\Theta)})^{-1} \widehat{\nabla_{\Gamma}\CL(\Gamma)}$ efficiently.
The target is just the solution of the following line system
\begin{equation}
	\label{eqn:linear_system}
	\begin{aligned}
			(\widehat{\nabla^2_{\Gamma}{\CL}(\Theta)})\bfy =  \widehat{\nabla_{\Gamma}\CL(\Gamma)}, 
	\end{aligned}
\end{equation}
where $\bfy\in \RB^d$. 
	
Conjugate gradient algorithm~\citep{nocedal2006numerical}, a well-studies algorithm in the context of numerical optimization, is suited to solve it.
	Note that in conjugate gradient algorithm, matrix-vector product $\widehat{\nabla^2_{\Gamma}{\CL}(\Theta)}\bfx$ (where $\bfx\in \RB^d$) is frequently calculated, thus when $S\ll d$, via making use of the special structure described in Equation~\eqref{eqn:monte_carlo_integration2}, 
	the computational complexity can be reduced to $O(Sd)$, instead of $O(d^2)$.
	
	The resulting algorithm is simple and listed in Algorithm~\ref{alg:I}, referred to as Second-Order Stochastic Gradient Variational Inference Scheme-I (SO-SGVI-I).
	\begin{algorithm}[]
		\caption{Scheme I}
		\label{alg:I}
		\begin{algorithmic}[1]
			\REQUIRE initial value $\Gamma_0$, 
			\ENSURE final value of variational parameter $\Gamma_t$.
			\FOR {$t = 1,2, \ldots$}
			\STATE Estimate gradient  $\widehat{\nabla_{\Gamma}\CL(\Gamma_{t-1})}$ according to Equation~\eqref{eqn:first_order_gradient_2}.
			\STATE Estimate Hessian matrix $\widehat{\nabla^2_{\Gamma}{\CL}(\Theta)}$ according to Equation~\eqref{eqn:monte_carlo_integration}. 
			\STATE Option I: Compute $(\widehat{\nabla^2_{\Gamma}{\CL}(\Theta)})^{-1}$ via iteratively using Sherman–Morrison formula, then update the variational parameter as $\Gamma_t = \Gamma_{t-1} - \epsilon (\widehat{\nabla^2_{\Gamma}{\CL}(\Theta)})^{-1}\widehat{\nabla_{\Gamma}\CL(\Gamma_{t-1})}$. 
			\STATE Option II: Solve the linear system $(\widehat{\nabla^2_{\Gamma}{\CL}(\Theta)})\bfy =  \widehat{\nabla_{\Gamma}\CL(\Gamma)}$ using conjugate algorithm. 
			Then update the variational parameter as $\Gamma_t = \Gamma_{t-1} -  \bfy$.
			\IF {Convergence condition is met}
			\STATE break.
			\ENDIF
			\ENDFOR
		\end{algorithmic}
	\end{algorithm}
	
\subsection{Scheme II}
\label{sec:schemeII}
\cite{agarwal2016second} devised a novel estimator for the Hessian matrix when optimizing the finite sums.
Here we adapt this strategy to variational inference setting.
The motivation is from the well known fact about the Taylor series expansion of the matrix inverse as follows.
\begin{lemma}
		\label{lemma:taylor}
		For a semi-positive definite matrix $\bfA\in\RB^{d\times d}$ satisfying that $\Vert\bfA\Vert< 1$\footnote{In this paper, $\Vert\cdot\Vert$ represents the spectral norm for a matrix and $l_2$ norm for a vector.}, we have that
		\begin{equation}
			\label{eqn:taylor}
			\begin{aligned}
				\bfA^{-1} = \sum_{i=0}^{\infty}(\bfI - \bfA)^i,
			\end{aligned}
		\end{equation}	
		where $\bfI$ denotes the identity matrix. 
	\end{lemma}
	Then a sequence of matrix $\{\bfB_i\}_{i=0}^{\infty}, \bfB_i\in\RB^{d\times d}$ is defined as
	\begin{equation}
		\begin{aligned}
			\bfB_i = \sum_{j=0}^{i}(\bfI - \bfA)^{j},
		\end{aligned}
	\end{equation}
	i.e., the first $i$ terms of Taylor expansion described Equation~\eqref{eqn:taylor}. 
	It is easy to find the limiting properties 
	\begin{equation}
		\begin{aligned}
			& \lim_{i\xrightarrow{}\infty} \bfB_i =  \bfA^{-1}
		\end{aligned}
	\end{equation}
	as long as the assumptions in Lemma~\ref{lemma:taylor} holds.
	Additionally, we have the recursion
	\begin{equation}
		\begin{aligned}
			& \bfB_i = \bfI +  (\bfI - \bfA)\bfB_{i-1}.\\
		\end{aligned}
	\end{equation}
	An unbiased estimator of Hessian matrix is devised based on this recursive formulation, given as
	\begin{equation}
		\label{eqn:recursion}
		\begin{aligned}
			& \widehat{\bfB_0}= \bfI, \\
			& \widehat{\bfB_i} = \bfI + (\bfI - \bfX_i)\widehat{\bfB_{i-1}}.
		\end{aligned}
	\end{equation} 
	where $\{\bfX_i \}$ are unbiased samples of the Hessian matrix $\nabla^2_{\Gamma}\CL(\Gamma)$. Concretely, 
	\begin{equation}
		\label{eqn:xi}
		\begin{aligned}
			\bfX_i = & \nabla_\Gamma \log q(\Theta^{(i)}\vert\Gamma)] [\nabla_\Gamma \log  q(\Theta^{(i)}\vert\Gamma)]^T \ln p(\Theta^{(i)},\bfX)\\
			& +  \nabla^2_\Gamma \log  q(\Theta^{(i)}\vert\Gamma)\ln p(\Theta^{(i)},\bfX),
		\end{aligned}
	\end{equation} 
	where $\Theta^{(1)}, \Theta^{(2)},\ldots$ are sampled i.i.d. from variational distribution $q(\Theta\vert\Gamma)$.
	Now we show that this estimator is unbiased for the inversion of Hessian matrix $\nabla^2_{\Gamma} \CL(\Gamma)$.
	\begin{lemma}
		$\widehat{\bfB_i}$ is an unbiased estimator for $\bfB_i$, i.e.,
		$\EB[\widehat{\bfB_i}] = \bfB_i$. Furthermore, we have
		\begin{equation}
		\label{eqn:estimator}
			\begin{aligned}
				\EB[\widehat{\bfB_i}]\xrightarrow{} ( \nabla^2_{\Gamma} \CL(\Gamma))^{-1}\ \ 
				\text{as}\ i\xrightarrow{}\infty.\\
			\end{aligned}
		\end{equation}
	\end{lemma}
	
	In practice, we attempt to avoid matrix-matrix\footnote{like direct computation of recursion in Equation~\eqref{eqn:recursion}} or matrix-vector product for computational efficiency. 
	Additionally, it is observed that $\bfX_i$ described in Equation~\eqref{eqn:xi} exhibits a special structure ((block) diagonal plus a rank-one correction) so that the computational complexity can be significantly reduced.
	The resulting algorithm is computationally light. 
	Because per-iteration computation involves only vector-vector product
	($O(d)$ complexity). 
	The computational complexity of each step is also listed. 
	Details are listed in Algorithm~\ref{alg:2}. 
	
	Note that in Lemma~\ref{lemma:taylor}, the assumption that $\Vert\bfA\Vert < 1$ is too restrictive for Hessian matrix, thus we circumvent this case via estimating the inverse of $\frac{1}{C_0}\bfA$ (corresponding to Step 6, 7, 12 in Algorithm~\ref{alg:2}), where constant $C_0$ is pre-specified and satisfies that $C_0 > \max\limits_{\Gamma} \Vert\nabla^2_{\Gamma}\CL(\Gamma)\Vert$. 
	In practice, $C_0$ usually takes a large value satisfying that $C_0 \gg \Vert\nabla^2_{\Gamma}\CL(\Gamma)\Vert$.	
	
\begin{algorithm}[]
\caption{Scheme II of Second-Order SVI}
\label{alg:2}
\begin{algorithmic}[1]
		\REQUIRE initial value $\Gamma_0$, maximal iteration $T_{\text{max}}$, tolerance $\eta$, constant $C_0 \gg \Vert\nabla^2_{\Gamma}\CL(\Gamma)\Vert$
		\ENSURE final value of variational parameter $\Gamma_t$.
		\FOR {$t = 1,2, \ldots$}
		\STATE Estimate gradient  $\widehat{\nabla_{\Gamma}\CL(\Gamma_{t-1})}$ according to Equation~\eqref{eqn:first_order_gradient_2}. \ \ \ \ \ \ \ \ \ \ \ \# $O(d)$
		\STATE $\bfy_0 = \widehat{\nabla_{\Gamma}\CL(\Gamma_{t-1})}$.
		\FOR {$j = 1,\ldots,T_{\text{max}}$}
		\STATE Sample $\Theta^{(j)}$ from the variational distribution $q(\cdot\vert\Gamma)$ and compute the Hessian matrix for $\Theta^{(j)}$, denoted $\bfX_i$. \ \ \ \ \ \ \ \ \ \ \ \# $O(d)$: Note that $\bfX_i$ is not computed explicitly here. 
		\STATE $\tilde{\bfX}_i = \bfX_i / C_0$.
		\STATE $\bfy_j = \widehat{\nabla_{\Gamma}\CL(\Gamma_{t-1})} + \frac{1}{C_0}\bfy_{j-1} - \tilde{\bfX}_i \bfy_{j-1}$.
		\STATE \# $O(d)$: $\tilde{\bfX}_i \bfy_{j-1}$ can be computed in linear time. 
		\IF {$\Vert\bfy_{j} - \bfy_{j-1}\Vert\leq \eta$}
		\STATE break.
		\ENDIF
		\ENDFOR	
	\STATE Then update the variational parameter as $\Gamma_t = \Gamma_{t-1} - (\frac{1}{C_0}\bfy_j)$. 
	\IF {Convergence condition is met}
	\STATE break.
	\ENDIF
	\ENDFOR
\end{algorithmic}
\end{algorithm}

	\bibliographystyle{unsrtnat}
	\bibliography{variational2}

\begin{thebibliography}{15}
\providecommand{\natexlab}[1]{#1}
\providecommand{\url}[1]{\texttt{#1}}
\expandafter\ifx\csname urlstyle\endcsname\relax
  \providecommand{\doi}[1]{doi: #1}\else
  \providecommand{\doi}{doi: \begingroup \urlstyle{rm}\Url}\fi

\bibitem[Robert and Casella(2013)]{robert2013monte}
Christian Robert and George Casella.
\newblock \emph{Monte Carlo statistical methods}.
\newblock Springer Science \& Business Media, 2013.

\bibitem[Jordan et~al.(1999)Jordan, Ghahramani, Jaakkola, and
  Saul]{jordan1999introduction}
Michael~I Jordan, Zoubin Ghahramani, Tommi~S Jaakkola, and Lawrence~K Saul.
\newblock An introduction to variational methods for graphical models.
\newblock \emph{Machine learning}, 37\penalty0 (2):\penalty0 183--233, 1999.

\bibitem[Chen et~al.(2015)Chen, Ding, and Carin]{chen2015convergence}
Changyou Chen, Nan Ding, and Lawrence Carin.
\newblock On the convergence of stochastic gradient mcmc algorithms with
  high-order integrators.
\newblock In \emph{Advances in Neural Information Processing Systems}, pages
  2278--2286, 2015.

\bibitem[Chen et~al.(2014)Chen, Fox, and Guestrin]{chen2014stochastic}
Tianqi Chen, Emily~B Fox, and Carlos Guestrin.
\newblock Stochastic gradient hamiltonian monte carlo.
\newblock \emph{arXiv preprint arXiv:1402.4102}, 2014.

\bibitem[Li et~al.(2020)Li, Glass, and Sun]{fu2021mimosa}
Xinhao Li, Lucas~M Glass, and Jimeng Sun.
\newblock {MIMOSA}: Multi-constraint molecule sampling for molecule
  optimization.
\newblock \emph{AAAI}, 2020.

\bibitem[Qian et~al.(2021)Qian, Xiao, Glass, and Sun]{fu2021probabilistic}
Cheng Qian, Cao Xiao, Lucas~M Glass, and Jimeng Sun.
\newblock Probabilistic and dynamic molecule-disease interaction modeling for
  drug discovery.
\newblock In \emph{Proceedings of the 27th ACM SIGKDD Conference on Knowledge
  Discovery \& Data Mining}, pages 404--414, 2021.

\bibitem[Huang et~al.(2020)Huang, Fu, Glass, Zitnik, Xiao, and
  Sun]{huang2020deeppurpose}
Kexin Huang, Tianfan Fu, Lucas~M Glass, Marinka Zitnik, Cao Xiao, and Jimeng
  Sun.
\newblock Deeppurpose: A deep learning library for drug-target interaction
  prediction.
\newblock \emph{Bioinformatics}, 2020.

\bibitem[Paisley et~al.(2012)Paisley, Blei, and Jordan]{paisley2012variational}
John Paisley, David Blei, and Michael Jordan.
\newblock Variational bayesian inference with stochastic search.
\newblock \emph{arXiv preprint arXiv:1206.6430}, 2012.

\bibitem[Ranganath et~al.(2014)Ranganath, Gerrish, and
  Blei]{ranganath2014black}
Rajesh Ranganath, Sean Gerrish, and David~M Blei.
\newblock Black box variational inference.
\newblock In \emph{AISTATS}, pages 814--822, 2014.

\bibitem[Ranganath et~al.(2013)Ranganath, Wang, Blei, and
  Xing]{ranganath2013adaptive}
Rajesh Ranganath, Chong Wang, David~M Blei, and Eric~P Xing.
\newblock An adaptive learning rate for stochastic variational inference.
\newblock In \emph{ICML (2)}, pages 298--306, 2013.

\bibitem[Wang et~al.(2013)Wang, Chen, Smola, and Xing]{wang2013variance}
Chong Wang, Xi~Chen, Alex~J Smola, and Eric~P Xing.
\newblock Variance reduction for stochastic gradient optimization.
\newblock In \emph{Advances in Neural Information Processing Systems}, pages
  181--189, 2013.

\bibitem[Nocedal and Wright(2006)]{nocedal2006numerical}
Jorge Nocedal and Stephen Wright.
\newblock \emph{Numerical optimization}.
\newblock Springer Science \& Business Media, 2006.

\bibitem[Fan et~al.(2015)Fan, Wang, Beck, Kwok, and Heller]{fan2015fast}
Kai Fan, Ziteng Wang, Jeff Beck, James Kwok, and Katherine~A Heller.
\newblock Fast second order stochastic backpropagation for variational
  inference.
\newblock In \emph{Advances in Neural Information Processing Systems}, pages
  1387--1395, 2015.

\bibitem[Glass et~al.(2020)]{fu2020alpha}
Lucas~M Glass et~al.
\newblock $\alpha$-mop: Molecule optimization with $\alpha$-divergence.
\newblock In \emph{2020 IEEE International Conference on Bioinformatics and
  Biomedicine (BIBM)}, pages 240--244. IEEE, 2020.

\bibitem[Agarwal et~al.(2016)Agarwal, Bullins, and Hazan]{agarwal2016second}
Naman Agarwal, Brian Bullins, and Elad Hazan.
\newblock Second order stochastic optimization in linear time.
\newblock \emph{arXiv preprint arXiv:1602.03943}, 2016.

\end{thebibliography}

\end{document}